\documentclass[10pt,twocolumn,letterpaper]{article}

\usepackage{iccv}
\usepackage{times}
\usepackage{epsfig}
\usepackage{graphicx}
\usepackage{amsmath}
\usepackage{amssymb}
\usepackage{mathtools}
\usepackage{booktabs} % for borders and merged ranges
\usepackage{soul}
\usepackage[table,xcdraw]{xcolor}
\usepackage{multirow}
\usepackage{float}

\usepackage[pagebackref=true,breaklinks=true,letterpaper=true,colorlinks,bookmarks=false]{hyperref}

\usepackage[capitalize]{cleveref}
\crefname{section}{Sec.}{Secs.}
\Crefname{section}{Section}{Sections}
\Crefname{table}{Table}{Tables}
\crefname{table}{Tab.}{Tabs.}

\iccvfinalcopy % *** Uncomment this line for the final submission

\ificcvfinal\fi

\begin{document}

%%%%%%%%% TITLE
\title{3D VR Sketch Guided 3D Shape Prototyping and Exploration}

\author{Ling Luo$^{1,2}$
\and
Pinaki Nath Chowdhury$^{1,2}$
\and
Tao Xiang$^{1,2}$
\and
Yi-Zhe Song$^{1,2}$
\and 
Yulia Gryaditskaya$^{1,3}$
\and\\
$^{1}$SketchX, CVSSP, University of Surrey, United Kingdom \\
$^{2}$iFlyTek-Surrey Joint Research Center on Artificial Intelligence 
\\
CCMLab, Surrey Institute for People-Centered AI and CVSSP, University
of Surrey, United Kingdom\\
}

\maketitle
\ificcvfinal\thispagestyle{empty}\fi

%%%%%%%%% ABSTRACT
\begin{abstract}
3D shape modeling is labor-intensive, time-consuming, and requires years of expertise.
To facilitate 3D shape modeling, we propose a 3D shape generation network that takes a 3D VR sketch as a condition.
We assume that sketches are created by novices without art training and aim to reconstruct geometrically realistic 3D shapes of a given category. To handle potential sketch ambiguity, our method creates multiple 3D shapes that align with the original sketch's structure. We carefully design our method, training the model step-by-step and leveraging multi-modal 3D shape representation to support training with limited training data.
To guarantee the realism of generated 3D shapes, we leverage the normalizing flow that models the distribution of the latent space of 3D shapes.
To encourage the fidelity of the generated 3D shapes to an input sketch, we propose a dedicated loss that we deploy at different stages of the training process.
The code is available at \url{https://github.com/Rowl1ng/3Dsketch2shape}.
\end{abstract}

%%%%%%%%% BODY TEXT

\section{Introduction}
\label{sec:intro}
The demand for convenient tools for 3D content creation constantly grows as the creation of virtual worlds becomes an integral part of various fields such as architecture and cinematography.
Recently, several works have demonstrated how text and image priors can be used to create 3D shapes \cite{sanghi2022clip,liu2022iss,lin2022magic3d,fu2022shapecrafter,li20223ddesigner}.
However, it is universally accepted that text is much less expressive or precise than a 2D freehand sketch in conveying spatial or geometric information \cite{yu2016sketch,chowdhury2022fs,sangkloy2022sketch}.
Therefore, many works focus on sketch-based modeling from 2D sketches \cite{olsen2009sketch, bonnici2019sketch, bhattacharjee2020survey, camba2022sketch, 2017_Lun, li2018robust, delanoy20183d, wang20203d, zhong2020towards, zhong2020deep, zhang2021sketch2model, sketch2mesh, kong2022diffusion} as a convenient tool for creating virtual 3D content.
Yet, 2D sketches are ambiguous, and depicting a complex 3D shape in 2D requires substantial sketching expertise.
As Virtual Reality (VR) headsets and associated technologies progress \cite{jerald2015vr, hu2020gaze, martin2022scangan360}, 
 more and more works consider 3D VR sketch as an input modality in the context of 3D modeling \cite{yu2022piecewise,yu2021scaffoldsketch,rosales2019surfacebrush} and retrieval \cite{li20153d,li2015kinectsbr,li2016shrec,ye20163d,luo2020towards,luo2021fine,luo2022structure}. Firstly, 3D VR sketches are drawn directly in 3D and therefore provide a more immersive and intuitive design experience.  Secondly, 3D VR sketches offer a natural way to convey volumetric shapes and spatial relationships.
 % , enabling more precise and comprehensive shape reconstruction. 
 Moreover, the use of 3D VR sketches aligns with advancements in virtual reality technology, making the process of sketching and designing more future-proof and adaptable to emerging technologies.

 \begin{figure}
    \includegraphics[width=\linewidth]{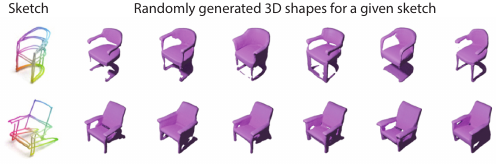}
  \caption{Given a VR (Virtual Reality) sketch input,  we generate 3D shape samples that satisfy three requirements: (1 - \emph{fidelity}) reconstructed shapes follow the overall structure of a quick VR sketch; (2 - \emph{diversity}) reconstructed shapes contain some diversity in shape details: such as a hollow or solid backrest, and (3 - \emph{realism}) reconstructions favor geometrically realistic 3D shapes of a given category.
  }
  \label{fig:teaser}
\end{figure}

Existing works on 3D shape modeling assume carefully created inputs and focus primarily on the interface and logistics of the sketching process.
In this paper, we introduce a novel method for 3D shape modeling that utilizes 3D VR sketching. Our method does not require professional sketch training or detailed sketches and is trained and tested on a dataset of sketches created by participants without art experience. 
This approach ensures that our model can accommodate a diverse range of users and handle sketches that might be less polished or precise, making it more accessible and practical for real-world applications.
Considering the sparsity of VR sketches, a single 3D shape model may not match the user's intention. Therefore, we advocate for generating multiple shape variations that closely resemble the input sketch, as demonstrated in \cref{fig:teaser}. 
The user can then either directly pick one of the models, or refine the design given the visualized 3D shapes, or multiple shapes can be used in some physical simulation process to select the optimal shape within the constraints of the VR sketch.

\begin{figure}[t]
  \centering
   \includegraphics[width=1.0\linewidth]{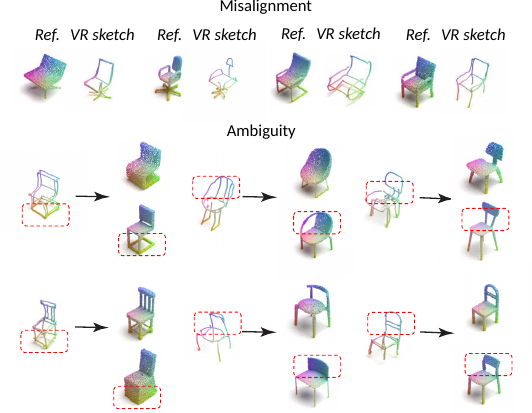}
   \caption{Example of misalignment and ambiguity of 3D sketch. Misalignment: the collected sketches and reference shapes have deviations in terms of the position and proportion of their parts. Ambiguity: due to the sparsity and abstract nature of sketches, strokes can be interpreted differently. For example, the strokes of a cube can represent either slender bars or a closed solid shape.
   }
   \label{fig:problems}
\end{figure}

Working with freehand VR sketches presents several challenges due to the lack of datasets.
We are aware of only one fine-grained dataset of VR sketches by Luo et al.~\cite{luo2021fine}, which we use in this work.
The challenge of working with this data comes from its limited size and the misalignment between sketches and 3D shapes. 
Luo et al.~\cite{luo2021fine} let participants sketch in an area different from the one where the reference 3D shape is displayed. 
This allows to model the scenario of sketching from memory or imagination,  however, results in a lack of alignment between 3D shapes and sketches, as shown in \cref{fig:problems}. 
Considering the misalignment of sketches and shapes in the dataset, and the ambiguity of the VR sketches, we aim to generate shapes with good fidelity to an input sketch, rather than the reference shape.

We represent our sketches as point clouds, and regress Signed Distance Fields (SDFs) values \cite{park2019deepsdf} representing 3D shapes.
Despite, the seemingly simple nature of the problem, we found that training an auto-encoder in an end-to-end manner results in poor performance due to a dataset's limited size and sketch-shape misalignments discussed above. We, therefore, start by training an SDF auto-decoder, similar to the one proposed by Park et al.~\cite{park2019deepsdf}. We then propose several losses that allow us to efficiently train our sketch encoder. 
In particular, we design a sketch fidelity loss, exploiting the fact that sketch strokes represent 3D shape surface points. Leveraging the properties of SDF, this implies that the regressed SDF values in the points sampled from sketch strokes should be close to zero.
To be able to sample multiple 3D shapes for a given input sketch, we adopt a conditional normalizing flow (CNF) model \cite{dinh2014nice}, trained to model the distribution of the latent space of 3D shapes. 
During the training of CNF, we again leverage the introduced sketch fidelity loss, improving the fidelity of reconstruction to the input sketch.

In summary, our contributions are the following: 
\begin{itemize}
    \item We, for the first time, study the problem of conditional 3D shape generation from rapid and sparse 3D VR sketches and carefully design our method to tackle the problem of (1) limited data, (2) misalignments between sketches and 3D shapes and (3) abstract nature of freehand sketches. 
    \item Taking into consideration the ambiguity of VR sketch interpretation, we design our method so that diverse 3D shapes can be generated that follow the structure of a given 3D VR sketch.  
\end{itemize}

\section{Related Work}
\subsection{Shape reconstruction and retrieval from 3D sketches}
Many earlier works targeted category-level retrieval \cite{li20153d,li2015kinectsbr,li2016shrec,ye20163d,luo2020towards}, but the field advancement stagnated due to the lack of fine-grained datasets. 
Recently several works addressed the problem of fine-grained retrieval from VR sketches \cite{luo2021fine,luo2022structure}, and introduced the first dataset which we use in this work. 
Luo et al.~\cite{luo2022structure} proposed a structure-aware retrieval that allows increasing relevance of the top retrieval results. 
However, retrieval methods are always limited to the existing shapes. 
Therefore, we explore conditional sketch generation, aiming to achieve good fidelity to the input, combined with generation results diversity.

Recently, Yu et al.~\cite{yu2022piecewise} considered the problem of VR sketch surfacing. 
Their work is optimization-based and assumes professional and detailed sketch input. 
Similarly to their work, we aim to achieve good fidelity of the reconstruction to an input sketch but take as input sparse and abstract sketches. Moreover, our approach is learning-based, which means we require a class shape prior, but we can handle abstract sketches, and inference is instant.

\subsection{3D shapes generation}

In addition to maximizing the fidelity to the input sketch, we aim to generate a set of shapes that are distinctive from each other. 
This diversity allows users to efficiently explore the design space of shapes resembling their initial sketch.

\subsubsection{Generative models}

Various 3D shape generative models have been proposed in the literature based on Generative Adversarial Networks (GANs) \cite{wu2016learning,achlioptas2018learning,chen2019learning,wu2020multimodal,zheng2022sdf}, 
Variational Auto-Decoders (VADs)\cite{park2019deepsdf,cheng2022cross}, 
autoregressive models \cite{sun2020pointgrow, wang2021sceneformer, mittal2022autosdf,yan2022shapeformer}, normalizing flow models \cite{sanghi2022clip} and more recently diffusion models \cite{kong2022diffusion,cheng2022sdfusion,nam20223d,zhou20213d,poole2022dreamfusion}. 
We chose to use normalizing flows trained in the latent space of our auto-encoder due to the simplicity of this model and the fact that it can be easily combined with any pretrained auto-encoder.
Recently, several concurrent works proposed the use of diffusion models in the latent space \cite{cheng2022sdfusion,nam20223d,poole2022dreamfusion}. 
Our network can easily be adapted to the usage of diffusion models instead of normalizing flows. 
The contribution of our work lies in the definition of the problem and the overall network architecture, as well as the introduction of appropriate loss functions and the training setup, allowing to generate multiple samples fitting the input 3D VR sketch, given limited training data.

\subsubsection{Conditional shape generation}
Similarly, diverse conditional generative models were considered that takes as input a sketch \cite{bhattacharjee2020survey, camba2022sketch, 2017_Lun, li2018robust, delanoy20183d, wang20203d, zhong2020towards, zhong2020deep, zhang2021sketch2model, sketch2mesh, kong2022diffusion, cheng2022cross}, an image \cite{han2019image, cheng2022sdfusion}, an incomplete scan in a form of a point cloud \cite{berger2017survey,wu2020multimodal,arora2021multimodal,zhou20213d,yan2022shapeformer}, a coarse voxel shape \cite{chen2021decor}, or a textual description \cite{cheng2022sdfusion,sanghi2022clip,fu2022shapecrafter,liu2022towards,poole2022dreamfusion}.

Sketch-/image-based reconstruction methods typically focus on the generation of only one output result for each input, while we aim at the generation of multiple 3D shapes.
Meanwhile, in the point cloud completion task, it is typically to infer the missing parts from the observed parts and generate multiple possible completion results. 
Their task, however, differs from our goal as  we do not want the network to synthesize non-existent parts, but only to create various shapes that match the sparse freehand sketch taking into account how humans might abstract 3D shapes. 
This is also the reason why the autoregressive approaches, such as  \cite{sun2020pointgrow,mittal2022autosdf,yan2022shapeformer}, are not suitable for our problem.

Text-guided 3D shape generation shares similar ambiguity properties as VR sketch-guided, i.e., diverse results may match the same input text. CLIP-Forge \cite{sanghi2022clip} employs pre-trained visual-textual embedding model CLIP to bridge text and 3D domains, and uses conditional normalizing flow to model the conditional distribution of latent shape representation given text or image embeddings. 
Zhengzhe et al. \cite{liu2022towards} introduce shape IMLE (Implicit
Maximum Likelihood Estimation) to boost results diversity while utilizing a cyclic loss to encourage consistency. 
In our work, we condition on a VR sketch rather than text and aim to obtain diverse 3D shapes that follow the input sketch structure.

\section{Method}

\begin{figure*}[t]
  \centering
   \includegraphics[width=\linewidth]{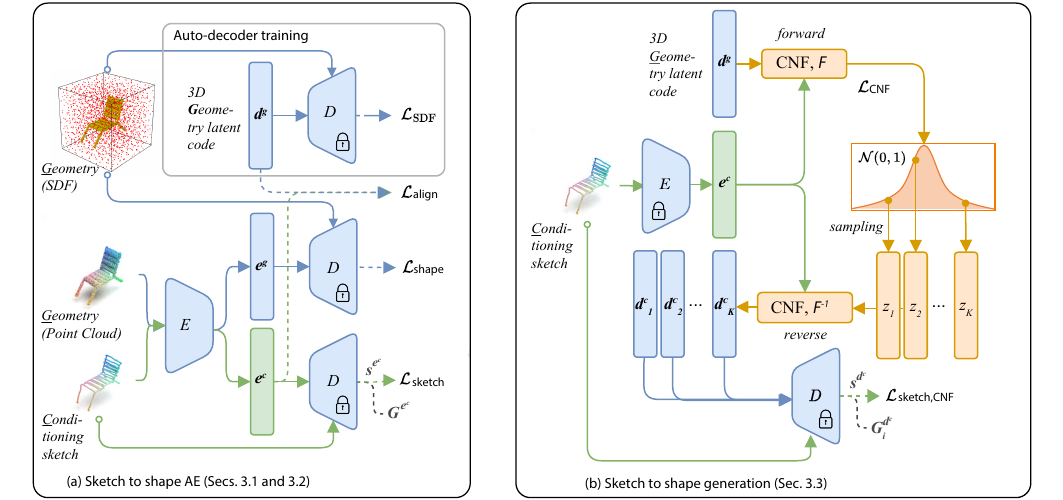}

   \caption{ Our method consists of 2 stages: (a) the first stage allows to obtain deterministic 3D shape reconstructions from input sketches, as described in \cref{sec:decoder,sec:encoder}, while (b) the second stage enables conditional 3D shape sample generation, as described in \cref{sec:generation}. 
   The auto-encoder (AE) is trained in three steps: first auto-decoder is trained, then the shape encoder is trained, and finally, the encoder is fine-tuned to jointly encode sketches and shapes.}
   \label{fig:network}
\end{figure*}

We present a conditional generation method that generates \emph{geometrically realistic} shapes of a specific category conditioned on abstract, sparse, and inaccurate freehand VR sketches. 
Our goal is to enforce the generation to stay close to the input sketch (\emph{sketch fidelity}) while providing sufficient \emph{diversity} of 3D reconstructions. 

The architecture of our method is shown in \cref{fig:network}. 
The method consists of two stages, where the first stage (\cref{fig:network} (a)) enables deterministic reconstruction for an input sketch, and the second stage allows for multiple sample generation (\cref{fig:network} (a)). 
We next describe the details of each stage.

\subsection{Shape decoder}\label{sec:decoder}
We represent 3D shapes using truncated signed distance functions (SDFs), as one of the most common 3D shape representations. This representation is limited to watertight meshes, but without loss of generality, here we assume that our meshes are watertight.

An SDF is a continuous function of the form:
\begin{equation}
    \mathrm{SDF}(x) = s \ : \ x \in \mathbb{R}^{3}, \ s \in \mathbb{R},
\end{equation}
where $x$ is a 3D point coordinates and $s$ is the signed distance to the closest shape surface (a negative/positive sign indicates that the point is inside/outside the surface). 
The underlying surface is implicitly represented by the iso-surface of $\mathrm{SDF}(\cdot)=0$, and can be reconstructed using marching cubes \cite{lorensen1987marching}.

Our goal is to reconstruct a 3D shape from a given VR sketch, however, we found that classical auto-encoder training frameworks on our problem perform poorly when trained in an end-to-end manner. 
This is caused by (1) a limited training set size, and (2) the fact that the sketches are not perfectly aligned with 3D shapes. 
Therefore, we first train a 3D shape auto-decoder, following Park et al. \cite{park2019deepsdf}.

\paragraph{Shape auto-decoder}\label{sec:shape_decoder}
The auto-decoder is trained by minimizing an $L_1$ 
loss between the ground truth and predicted truncated signed distance values. The decoder 
\begin{equation}
    \mathrm{D}_{\theta}([\boldsymbol{d}^g, p_{i}]) = \tilde{s}_i, \ p_{i} \in \mathbb{R}^{3}, \tilde{s}_i \in \mathbb{R}
\end{equation}
takes as input the 3D shape latent code $\boldsymbol{d}^g$ and the 3D point coordinates $p_{i}$; $[\cdot, \cdot]$ represents a concatenation operation. The decoder predicts the per point signed distance value $\tilde{s}_i$. 
Once the decoder is trained, we freeze its parameters $\theta$.

At inference time, we estimate the 3D shape latent code via Maximum-a-Posterior estimation as follows: 
\begin{equation}
\hat{\boldsymbol{{d}}^g}=\underset{\boldsymbol{d}^g}{\arg \min } \sum_{\left({p}_{i}, {s}_{i}\right) \in G} \mathcal{L}\left(D_\theta\left(\boldsymbol{d}^g, {p}_{i}\right), s_i\right)+\frac{1}{\sigma^2}\|\boldsymbol{d}^g\|_2^2
\label{eq:inverse}
\end{equation}
where the latent vector $\boldsymbol{d}^g$ is initialized randomly, and the sum is taken over points of the 3D shape geometry $G$.
We treat the estimated latent vector $\hat{\boldsymbol{{d}}^g}$ as a ground truth shape embedding and denote it as $\boldsymbol{d}^g$ for simplicity.

\subsection{Encoding VR sketches.}\label{sec:encoder}
We then train a sketch encoder that maps sketches to the latent space of our 3D shape decoder. 

Due to the sparsity of sketch inputs, we represent them as point clouds. 
We observe that if we only use sketches for training, we obtain very poor generalization to the test set. 
Therefore, we exploit a joint encoder for sketches and 3D shapes, using  a PointNet++ \cite{qi2017pointnet++} encoder.  
The encoder $\mathrm{E}_{\phi}(\cdot): \mathbb{R}^{N_{s} \times 3} \rightarrow \mathbb{R}^{256}$ embeds randomly sampled points from input shape surface $G \in \mathbb{R}^{N_{s} \times 3}$ or sketch strokes $C \in \mathbb{R}^{N_{s} \times 3}$ to a feature vector $\boldsymbol{e}^{g} \in \mathbb{R}^{256}$ and $\boldsymbol{e}^{c}\in \mathbb{R}^{256}$, respectively. 

The encoder parameters $\{\phi\}$ are optimized by several losses at training time:
\begin{equation}
    \mathcal{L} = \mathcal{L}_{\text{shape}} + \mathcal{L}_{\text{sketch}} + \mathcal{L}_{\text{align}},
    \label{eq:full_loss_emncoder}
\end{equation}
where $\mathcal{L}_{\text{shape}}$ ensures that we can accurately regress 3D shapes SDF from a 3D shape point cloud representation, $\mathcal{L}_{\text{sketch}}$ ensures that the reconstructed 3D shape is close to the input sketch, and $\mathcal{L}_{\text{align}}$ establishes a connection between sparse VR sketches and 3D shapes. 

\subsubsection{3D shape auto-encoder}
First loss, $\mathcal{L}_{\text{shape}}$, operates only on 3D shape inputs, and ensures that the full network $\mathrm{D}_{\theta}([\mathrm{E}_{\phi}(\cdot), p_{i}])$ functions as a 3D shape autoencoder. 
First, we minimize the sum of $L_1$ losses between the truncated predicted and ground truth SDF values of sampled points $p_{i} \in \mathbb{R}^{3}$:
\begin{equation}\label{eq:SDF_loss}
    \mathcal{L}_{\text{SDF}}(\phi) = \frac{1}{N_{s}} \sum_{p_{i}} | \text{tr}(\mathrm{D}_{\theta}([\boldsymbol{e}^{g}, p_{i}])) - \text{tr}(s(p_{i}))|
\end{equation}
where the decoder parameters $\theta$ are from our auto-decoder and are frozen when training the sketch/shape encoder; $s(\cdot)$ denotes ground-truth 3D shape SDF values, and $\text{tr}(\cdot) \triangleq \min(\delta,\max(-\delta, \cdot))$. 

Additionally, we ensure that the 3D shape embedding $\boldsymbol{e}^{g}=\mathrm{E}_{\phi}(f)$ maps directly to a latent representation of a 3D shape $\boldsymbol{d}^g$, computed with \cref{eq:inverse}:
\begin{equation}
\mathcal{L}_{g\text{-}L_1}(\phi)=|\boldsymbol{e}^{g}-\boldsymbol{d}^g|.
\end{equation}

We also found that adding contrastive latent term loss increases performance. 
Let's assume that our mini-batch consists of $N_g$ shapes. 
First, we obtain latent representations $\{\boldsymbol{d}^g_1,\dots,\boldsymbol{d}^g_{N_g}\}$ of 3D shapes, using \cref{eq:inverse}.
Then, we formulate our contrastive loss term as follows:
\begin{equation}
\mathcal{L}_{\text{g-NCE}}(\phi)=-\sum_{i=1}^{N_g}\left[\log \frac{\exp \left(-\left|\boldsymbol{e}^{g}_{i}-\boldsymbol{d}^{g}_{i}\right|\right)}{\sum_{j=1}^{N_g} \exp \left(-\left|\boldsymbol{e}^{g}_{i}-\boldsymbol{d}^g_{j} \right|\right)}\right].
\label{eq:f-NCE}
\end{equation}
Therefore, the shape loss $\mathcal{L}_{\text{shape}}$ is the sum of the three losses defined above:
\begin{equation}
\mathcal{L}_{\text{shape}} = \mathcal{L}_{\text{SDF}}(\phi) + \mathcal{L}_{g\text{-}L_1}(\phi) + \mathcal{L}_{g\text{-NCE}}(\phi).
\label{eq:L_shape}
\end{equation}

\subsubsection{Sketch loss}

Given the misalignment between sketches and reference 3D shapes in the dataset \cite{luo2020towards}, as we show in \cref{fig:problems}, we aim for the reconstruction result to stay close to the sketch input. 
To achieve this goal, we design a sketch loss $\mathcal{L}_{\text{sketch}}$. 

Since a sketch is a sparse representation of a 3D shape, 
the intended 3D shape surface should lie in the vicinity of sketch stroke points.
Therefore, the reconstructed SDF values at those points should be close to zero. Formally, we define this loss as follows:
\begin{equation}
 \mathcal{L}_{\text{sketch}}(\phi) = \frac{1}{N_{s}}\sum_{i= 1}^{N_s}|D(\boldsymbol{e}^{c},p^c_{i})|,
 \label{eq:sketch_sdf_loss}
\end{equation}
where $p^c_{i}\in C$ is the $i$-th sample points from the conditioning sketch, and $N_s$ is the number of sampled points in a sketch.

\subsubsection{Sketch-shape latent space alignment}
The considered so far losses do not explicitly ensure that there is a meaningful mapping between sketches and 3D shapes' latent representations.
Therefore, we design additional losses encouraging an alignment in the feature space.

First, we introduce a contrastive loss, similar to the one in \cref{eq:f-NCE}, leveraging that sketches in our dataset contain reference shapes. It takes the following form:
\begin{equation}
\mathcal{L}_{c\text{-NCE}}(\phi)=-\sum_{i=1}^{N_c}\left[\log \frac{\exp \left(-\left|\boldsymbol{e}^c_{i}-\boldsymbol{d}^g_i\right|\right)}{\sum_{j=1}^{N_g} \exp \left(-\left|\boldsymbol{e}^c_{i}-\boldsymbol{d}^g_j\right|\right)}\right],
\label{eq:f-NCE}
\end{equation}
where $N_c$ is the number of sketches and $N_g$ is the number of shapes in the mini-batch. 
This loss pulls the encodings of a sketch and a reference shape closer than the encodings of a sketch and non-matching 3D shapes.

Additionally, we minimize the $L_1$ distance between the sketch $C$ embedding $\boldsymbol{e}^{c}=E_\phi(C)$ to the \emph{ground-truth} shape latent code $\boldsymbol{d}^g$:
\begin{equation}
\mathcal{L}_{c\text{-}L_1}(\phi)=|\boldsymbol{e}^{c} - \boldsymbol{d}^g|.
\end{equation}

Finally, the alignment loss is the sum of the two losses:
\begin{equation}
    \mathcal{L}_{\text{align}} = \mathcal{L}_{c\text{-}L_1} + \mathcal{L}_{c\text{-NCE}} 
\end{equation}

\subsection{Conditional shape generation}\label{sec:generation}

As shown in \cref{fig:problems}, a sparse sketch can represent multiple 3D shapes, generally following the sparse sketch strokes. 
Therefore, we would like to be able to generate multiple 3D shapes given a sketch. We achieve this by training a conditional normalizing flow (CNF) model in the latent space. 

Specifically, we model the conditional distribution of shape embeddings using a RealNVP network \cite{dinh2016density} with five layers as in \cite{sanghi2022clip}. 
It transforms the probability distribution of shape feature embedding $p_{\boldsymbol{d}}(\boldsymbol{d}^g)$ to a unit Gaussian distribution $p_z(\boldsymbol{z})$. 
We obtain the sketch embedding vector $\boldsymbol{e}^{c}$ as described in the previous section, which serves as a condition for our normalizing flow model. 
Please note that the sketch encoder parameters $\phi$ are frozen at this stage.
The sketch condition $\boldsymbol{e}^{c}$ is concatenated with the matching 3D shape feature vector $\boldsymbol{d}^g$ at each scale and translation coupling layers, following RealNVP \cite{dinh2016density}:

\begin{gather}
\quad {z}^{1: d}={{d}^g}^{1:d} \quad \text{and} \\ 
{z}^{d+1: D}={{d}^g}^{d+1: D} \odot \exp \left(s\left(\left[\boldsymbol{e}^{c} ; {{d}^g}^{1: d}\right]\right)\right)+t\left(\left[\boldsymbol{e}^{c} ; {{d}^g}^{1: d}\right]\right)
\end{gather}
where $s(\cdot)$ and $t(\cdot)$ are the scale and translation functions, parameterized by a neural network, as described in \cite{dinh2016density}.

The idea of the normalizing flow model is to approximate a complicated probability distribution with a simple distribution through a sequence of invertible nonlinear transforms. 
We train the flow model by maximizing the log-likelihood $\log \left(p_{\boldsymbol{d}}\left({\boldsymbol{d}}\right)\right)$:

\begin{equation}
\label{eq:nf_loss}
\begin{split}
\mathcal{L}_{\text{CNF}} & = -\log \left(p_{\boldsymbol{d}}\left(\boldsymbol{d}\right)\right) \\ 
& =-(\log \left(p_z\left({z}\right)\right)+\log \left(\left|\operatorname{det}\left(\frac{\partial F\left(\boldsymbol{d}\right)}{\partial {z}^T}\right)\right|\right)),
\end{split}
\end{equation}
where $F(\cdot)$ is the normalizing flow model, and ${\partial F\left(\boldsymbol{d}\right)}/{\partial {z}^T}$ is the Jacobian of $F$ at $\boldsymbol{d}$. 

\paragraph{Sketch fidelity.} 
To ensure the fidelity of the generated 3D shapes to an input sketch, we additionally train the flow model with a loss similar to an $\mathcal{L}_{sketch}$ loss (\cref{eq:sketch_sdf_loss}). 

First, the flow module $F$ is updated with the gradients from $\mathcal{L}_{\text{CNF}}$.
Then, for each sketch condition $\boldsymbol{e}^{c}$, we randomly sample $K$ different noise vectors $\{z_k\}$ from the unit Gaussian distribution $z_k\in\mathcal{N}(0,1)$, as shown in \cref{fig:network} (b). 
These noise vectors are mapped back to the shape embedding space through the reverse path of the flow model. 
During training, the obtained shape embeddings $\{\boldsymbol{d}^c_k\}$ are fed to the implicit decoder $D_\theta(\cdot)$ together with sketch stroke points $p^c_{i}$. Formally, this loss takes the following form: 
\begin{equation}
\label{eq:sketch_sdf_loss_cne}
 \mathcal{L}_{\text{sketch,CNF}} = \frac{\lambda}{N_{s}K}\sum_{i= 1}^{N_s}\sum_{k= 1}^{K}|D\left(\boldsymbol{d}^c_k,p^c_{i}\right)|,
\end{equation}
where $N_{s}$ is a number of sketch stroke points, as before, and $\lambda$ is a hyper-parameter set to $100$ to increase the relative importance of this loss. 
In each mini-batch, we first propagate gradients from $\mathcal{L}_{\text{CNF}}$, and then from $\mathcal{L}_{\text{sketch,CNF}}$.

\paragraph{Conditional shape generation.}
During inference, given an input sketch, represented as a set of points, we first obtain its embedding $\boldsymbol{e}^{c}$ using the encoder $E_\phi(\cdot)$. 
We then condition the normalizing flow network with $\boldsymbol{e}^{c}$ and a random noise vector sampled from the unit Gaussian distribution to obtain a shape embedding $\boldsymbol{d}^c$. 
We obtain the mean embedding by using the mean of the normal distribution. 
Finally, this shape embedding is fed into implicit decoder $D_\theta(\cdot)$ to obtain a new set of SDF values $\{s^{d^c} | s^{d^c} = D(\boldsymbol{d}^c)\}$. 
A 3D geometry is then reconstructed by applying the marching cubes algorithm \cite{lorensen1987marching}.

\section{Experiments}

\subsection{Implementation Details}
\label{sec:implementation_details}
\paragraph{Auto-decoder}
We train a decoder, similar to \cite{park2019deepsdf}, to regress the continuous SDF value for a given 3D space point and latent space feature vector. 
Our decoder $\mathrm{D}_{\theta}: \mathbb{R}^{(256+3)} \rightarrow \mathbb{R}$ consists of $5$ feed-forward layers, each with dropouts. All internal layers are $512$-dimensional and have ReLU non-linearities.
The output layer uses {tanh} non-linearity to directly regress the continuous SDF scalar values.
Similar to \cite{park2019deepsdf}, we found training with batch normalization to be unstable and applied the weight-normalization technique. 

During training, for each shape, we sample locations of the 3D points at which we calculate SDF values. We sample two sets of points: close to the shape surface and uniformly sampled in the unit box. 
Then, the loss is evaluated on the random subsets of those pre-computed points. 
During inference, the 3D points are sampled on a regular ($256 \times 256 \times 256$) grid.

\paragraph{Encoder and Normalizing flow}
We train with an Adam optimizer, where for the encoder training the learning rate is set to $1e-3$, and for the normalizing flow model, it is set to $1e-5$.
Training is done on 2 Nvidia A100 GPUs.

When training a sketch encoder jointly on sketches and shapes each mini-batch consists of 12 sketch-shape pairs and additional 24 shapes that do not have a paired sketch.
When training CNF model, each mini-batch consists of 12 sketch-shape pairs.

We train the encoder and the conditional normalizing flow for 300 epochs each.
The encoder is however trained in two steps. 
First, it is pre-trained using 3D shapes only, using $\mathcal{L}_{\text{shape}}$ loss, defined in \cref{eq:L_shape}.  
The performance of the shape reconstruction from this step is provided for reference in the 1st line in \cref{tab:encoder}. 
The encoder is then fine-tuned using sketches and shapes with the full loss given by \cref{eq:full_loss_emncoder}. 

To train the sketch/shape encoder we sample $N_s=4096$ points from sketch strokes and shape surface, respectively.
Please refer to the supplemental for additional details.

\subsection{Datasets} 
For training and testing, we use the only available fine-grained dataset of freehand VR sketches by Luo et al.~\cite{luo2021fine}\footnote{\url{https://cvssp.org/data/VRChairSketch/}}. 
The dataset consists of 1,005 sketch shape pairs for the chair category of ShapeNet \cite{shapenet}.
We follow their split to training and test sets, containing 803 and 202 shape-sketch pairs, respectively. 
The 6,576 shapes from the ShapeNetCore-v2, non-overlapping with the 202 shapes in the test set, are used for training the auto-decoder and sketch/shape encoder.

\paragraph{Alignment of multiple data types:} 
The sketches in the used dataset have a consistent orientation with reference 3D shapes, but might be not well aligned horizontally and vertically to the references, and can have a different scale.
We sample shape point clouds and compute SDF values for the normalized 3D shapes as provided in ShapeNetCore-v2, which ensures consistency between the two 3D shape representations. 
We then normalize the sketches to fit a unit bounding box, following the normalization in the ShapeNetCore-v2. 
To further improve alignment between sketches and 3D shapes, we translate sketches, so that their centroids match the centroids of reference shapes.

\subsection{Evaluation Metrics} 

Following prior work, we choose a bidirectional Chamfer distance (CD) as the similarity metric between two 3D shapes. CD measures the average shortest distance from one set of points to another. 
To compute CD, we randomly sample 4,096 points from 3D meshes.

\paragraph{Shape fidelity, $\mathcal{F}_{\text{shape}}(\cdot)$}
First, we evaluate the ability of our auto-encoder to faithfully regress 3D shape SDF values given a 3D shape point cloud.
We evaluate the fidelity of the regressed 3D shape, $G^{e^g}$, to the ground-truth 3D shape, $G$, as follows:  $\mathcal{F}_{\text{shape}}(G^{e^g}) = CD(G,G^{e^g})$. 

Then, while the sketches in the used dataset do not align perfectly with reference 3D shapes and contain ambiguity, it is meaningful to expect that the reconstructed 3D shape still should be close to the reference 3D shape. 
Therefore, we evaluate how close the reconstructed 3D shapes are to the ground-truth when 
(1) the shape is reconstructed from the sketch embedding $\boldsymbol{e}^c$, denoted as $G^{{e}^c}$; 
(2) the shape is reconstructed from the predicted conditional mean $\bar{z}^c$ of the CNF model, denoted as $G^{\bar{z}^c}= D(F^{-1}(\bar{z}^c))$; and 
(3) the shape is reconstructed from a random sample from the latent space of the CNF model, denoted as $G^{d^c}$. 
The respective losses are: $\mathcal{F}_{\text{shape}}(G^{{e}^c})$, $\mathcal{F}_{\text{shape}}(G^{\bar{z}^c})$ and 
$\mathcal{F}^{\text{avg}}_{\text{shape}}(G^{d^c})=\sum_{i=1}^5\mathcal{F}_{\text{shape}}(G^{d^c}_{i})$, where in the latter case we generate $5$ samples and report an average loss value.

\paragraph{Sketch fidelity, $\mathcal{F}_{\text{sketch}}(\cdot)$}
Since the used sketches are ambiguous and are not perfectly aligned to a reference, we evaluate the fidelity of the reconstructions to the sketch input, using the loss similar to \cref{eq:sketch_sdf_loss}:
$\mathcal{F}_{\text{sketch}}(G^c) = \frac{1}{N_{s}}\sum_{i= 1}^{N_s}s^c(p^c_i)$,
where $p^c_i$ is the $i$-th sample point from the input sketch and $s^c$ denotes the predicted SDF. 
With that, we define $\mathcal{F}^{avg}_{\text{sketch}}(s^{d^c}) = \frac{1}{5}\sum_{i=1}^5 \mathcal{F}_{\text{sketch}}(s^{d^c}_{i})$, as the average fidelity of multiple samples from the CNF model space to an input sketch. 

\paragraph{Diversity, $\mathcal{D}_{\text{gnrtns}}$} 
To measure the diversity of the generated shapes, we formulate the pair-wise similarity of generated shapes using CD. Specifically, for any two generated shapes conditioned on the same sketch, we compute their CD, and finally report the mean of all pairs, which we refer to as $\mathcal{D}_{\text{gnrtns}}$.

\subsection{Results}
\begin{figure*}[h]
  \centering
   \includegraphics[width=\linewidth]{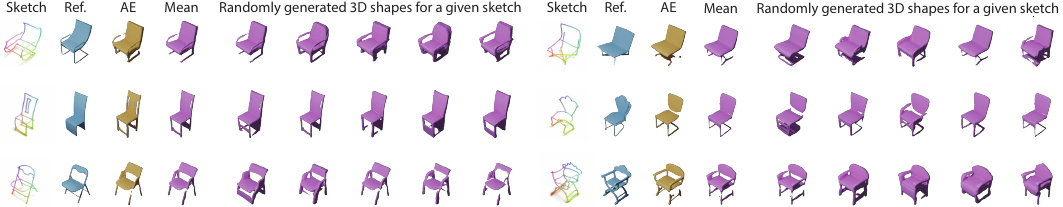}

   \caption{Generation results. `Ref.' shows the reference 3D shape. `AE' shows the deterministic prediction by our AE from the first stage of our method. `Mean' denotes the shape reconstructed from the sample corresponding to the mean of the conditional distribution. And finally, we  show 5 randomly generated shapes conditioned on the input sketch, sorted in the order of fidelity to a reference shape.
    }
   \label{fig:main}
\end{figure*}

We first evaluate the reconstruction performance of our AE and then evaluate multiple shape generation, conditioned on the input sketch. \cref{fig:main} shows qualitative results for both stages.

\subsubsection{Deterministic sketch to shape generation}

\begin{table}[t]\centering
\resizebox{\columnwidth}{!}{%
\begin{tabular}{ll|cc}\toprule
Method &Loss &$\mathcal{F}_{shape}(G^{{e}^c}) \downarrow$ &$\mathcal{F}_{shape}(G^{e^g}) \downarrow$ \\\midrule
Shape AE &$\mathcal{L}_{\text{shape}}$ &0.834 &\textbf{0.110} \\\midrule
\multirow{2}{*}{Sketch AE} &$\mathcal{L}_{\text{SDF}}$ &\ul{0.437} &0.581 \\
&$\mathcal{L}_{g\text{-}L_1} + \mathcal{L}_{\text{sketch}}$ &0.504 &0.321 \\\midrule
\textbf{Joint AE} &$\mathcal{L}$ &\textbf{0.357} &\ul{0.126} \\
\bottomrule
\end{tabular}
}
\vspace{1pt}
\caption{Evaluation of auto-encoder training strategies with respect to the fidelity ($\mathcal{F}_{shape}(\cdot)$) of the reconstructed 3D shapes to the reference/ground-truth 3D shapes, depending on the used data.
Here, $G^{{e}^c}$ and $G^{e^g}$ are reconstructions from an input sketch and 3D shape, respectively. 
Shape AE, Sketch AE, and Joint AE stand for training encoder only with shape inputs, sketch inputs, or both, respectively.
}
\label{tab:encoder}
\end{table}

Our first goal is to learn to map sketches to 3D shapes in a deterministic fashion. 
One of the challenges in our work comes from the limited dataset size, which is a common factor that should be taken into consideration when working with freehand sketches.
Therefore, we proposed training the sketch-to-shape auto-encoder in multiple steps, and in addition, we propose to use a joint auto-encoder, and we use 3D shapes without paired sketches in the NCE loss to improve robustness.
\cref{tab:encoder} shows that our strategy indeed outperforms alternative strategies. It allows reconstructing 3D shapes similar to reference 3D shapes, as shown by $\mathcal{F}_{shape}(G^{{e}^c})$. The fact that $\mathcal{F}_{shape}(G^{e^g})$ stays low in our proposed design implies that if the sketch is very detailed and accurate, we will obtain careful 3D shape reconstructions.

\begin{table}[t]\centering
\resizebox{\columnwidth}{!}{%
\begin{tabular}{l|cc}\toprule
Method &$\mathcal{F}_{shape}(G^{{e}^c}) \downarrow$ &$\mathcal{F}_{shape}(G^{e^g}) \downarrow$ \\\midrule
$\mathcal{L}_{\text{L1}}$ &0.418 &0.199 \\
$\mathcal{L}_{\text{L1}} + \mathcal{L}_{\text{SDF}}$ &0.374 &\ul{0.140} \\
$\mathcal{L}_{\text{L1}} + \mathcal{L}_{\text{SDF}} + \mathcal{L}_{\text{NCE}}$ &\ul{0.373} &\textbf{0.126} \\
\midrule
$\mathcal{L}_{\text{L1}} + \mathcal{L}_{\text{SDF}} + \mathcal{L}_{\text{NCE}} + \mathcal{L}_{\text{sketch}}$ &\textbf{0.357} &\textbf{0.126} \\
\bottomrule
\end{tabular}
}
\vspace{1pt}
\caption{Evaluation of auto-encoder training strategies with respect to the fidelity ($\mathcal{F}_{shape}(\cdot)$) of the reconstructed 3D shapes to the reference/ground-truth 3D shapes, depending on the used loss function.
Here, we group together sketch and shape $L_1$ and $NCE$ losses.}
\label{tab: encoder_loss}
\end{table}

\cref{tab: encoder_loss} demonstrates the importance of individual loss terms. 
It shows that the shape reconstruction loss $\mathcal{L}_{\text{SDF}}$ ensures that we can reconstruct shapes well when the input is dense (the case for the shape point cloud or very detailed sketches).
The sketch fidelity loss $\mathcal{L}_{\text{sketch}}$ ensures that the reconstructed shape is following the structure of an input sketch. 
Finally, NCE losses improve both the sketch and shape fidelity criteria of the reconstructed results. 

\subsubsection{Conditional sketch to shape generation}

\begin{table*}[t]\centering
  \normalsize
\begin{tabular}{lc|c|ccc}\toprule
Method &$K$ &$\mathcal{F}_{shape}(G^{{e}^c}) \downarrow$ &$\mathcal{F}^{avg}_{\text{sketch}}(G^{d^c}) \downarrow$ &$\mathcal{F}^{\text{avg}}_{\text{shape}}(G^{d^c}) \downarrow$ &$\mathcal{D}_{\text{gnrtns}} \uparrow$ \\\midrule
Joint AE &- &\textbf{0.357} &- &- &- \\\midrule
CNF($\boldsymbol{e}^g$) &- &0.373 &0.026$\pm$0.036 &\textbf{0.380} &0.043 \\
CNF($\boldsymbol{d}^g$) &- &0.385 &0.030$\pm$0.039 &\ul{0.420} &\textbf{0.165} \\\midrule
\multirow{3}{*}{CNF($\boldsymbol{d}^g$) + $\mathcal{L}_{\text{sketch,CNF}}$} &1 &\ul{0.366} &0.019$\pm$0.034 &0.422 &0.158 \\
&4 &0.397 &\ul{0.018$\pm$0.034} &0.448 &\textbf{0.165} \\
&8 &0.368 &\textbf{0.017$\pm$0.034} &0.431 &\ul{0.161} \\
\bottomrule
\end{tabular}
\vspace{1pt}
  \caption{Ablation of design choices for the CNF model. `Joint AE' stands for the result of our autoencoder model, and provides the estimated fidelity of the deterministic reconstruction to a sketch reference 3D shape. 
  `CNF' stands for a conditional normalizing flow. 
  $K$ refers to the number of samples used to compute $\mathcal{L}_{\text{sketch,CNF}}$ during training. 
  $\mathcal{F}^{avg}_{\text{sketch}}$ measures the average fidelity of the reconstructed 3D shape samples to an input sketch. 
  $\mathcal{F}^{\text{avg}}_{\text{shape}}$ measures the average fidelity of the reconstructed 3D shape samples to a reference shape. 
      $\mathcal{D}_{\text{gnrtns}}$ measures the diversity of the the reconstructed 3D shape samples.
  All fidelity measures are multiplied by $1e2$.}
  \label{tab:generation}
\end{table*}

Next, we conduct a number of experiments to assess the proposed conditional generation framework. 

\paragraph{Shape encoding choices}
Note that when training CNF model, we use the shape latent code, $\boldsymbol{d}^g$, obtained via an inversion process with \cref{eq:inverse}.
\cref{tab:generation}, lines 2 and 3, shows that this allows to greatly increase the diversity of the generated results compared to using latent shape codes, $\boldsymbol{e}^g$, obtained from the encoder. This comes with a small decrease in fidelity to the reference shape, while the fidelity to the sketch increases a little bit.  
This result reinforces our design choice of training the auto-decoder first, providing a richer latent space.

\begin{figure}[th]
  \centering
   \includegraphics[width=\linewidth]{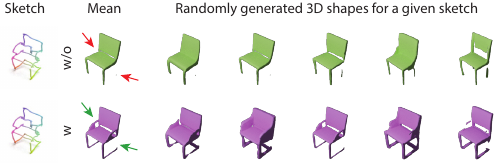}
   \caption{Comparison of the generated samples conditioned on the input sketch when $\mathcal{L}_{\text{sketch, CNF}}$ is used (purple) or not (green).
   This example shows that the sketch fidelity loss indeed results in better fidelity to a sketch input: all generated shapes when the loss is used contain handles and better respect the shape of chair legs/support. `Mean' denotes the shape reconstructed from the sample corresponding to the mean of the conditional distribution.}
   \label{fig:CNF_sketch_loss}
\end{figure}

\paragraph{Sketch fidelity loss in CNF model}
\cref{tab:generation}, lines 3 and 4, show that sketch consistency loss $\mathcal{L}_{\text{sketch, CNF}}$ results in much better sketch fidelity while maintaining comparable diversity. 
Varying the number of samples $K$ from the CNF latent space, we can further adjust the balance between sketch fidelity and diversity (\cref{tab:generation}, lines 4-6). 
We use the model with $K=8$ samples for the visual results in all our figures.
The advantage of this loss is demonstrated visually in \cref{fig:CNF_sketch_loss}.
It can be observed that the proposed loss encourages the network to always reconstruct some shape structure near the sketch strokes.

\subsection{Comparison to retrieval}
\begin{figure}[t]
  \centering
   \includegraphics[width=\linewidth]{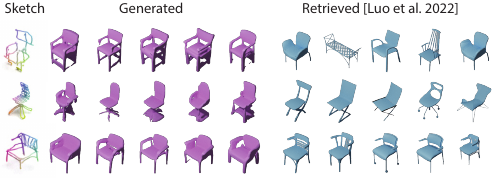}
   \caption{Comparison to the retrieval results by Luo et al.~\cite{luo2022structure}.}
   \label{fig:retrieval}
\end{figure}

\cref{fig:retrieval} shows comparison to the retrieval results by the state-of-the-art method \cite{luo2022structure} that is designed to retrieve structurally-similar shapes. It can be observed that generation can be more robust to shapes that are not common shapes in a 3D shape gallery. However, the reconstruction quality of our method is limited, and some shapes still do not look like real-world shapes, missing details. 

\section{Conclusion and Discussion}

We present the first method for multiple 3D shape generation conditioned on sparse and abstract sketches. 
We achieve good fidelity to the input sketch combined with the shape diversity of the generated results. 
In our work, we show how to efficiently overcome the limitation of small datasets. Our experiments are currently limited to a single category, but none of the components of our method explicitly exploits any priors about this category.
In the future, we would like to extend this work by (1) further improving the input sketch fidelity, potentially taking perceptual multi-view losses into account during training, and (2) considering alternative shape representation for our auto-decoder. 

{\small
\bibliographystyle{ieee_fullname}
\bibliography{egbib}
}

\newpage 

\renewcommand\thefigure{S\arabic{figure}}
\renewcommand\thetable{S\arabic{table}}
\renewcommand\thesection{S\arabic{section}}
\renewcommand\thesubsection{S\arabic{section}.\arabic{subsection}}

\setcounter{section}{0}
\section{Additional results}

\begin{figure}[th]
  \centering
   \includegraphics[width=\linewidth]{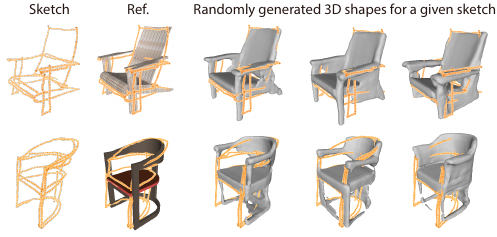}
   \caption{Alignment between sketches and shapes visualized using MeshLab's snapshot.}
   \label{fig:alignment}
\end{figure}

\begin{figure*}[h]
  \centering
   \includegraphics[width=\linewidth]{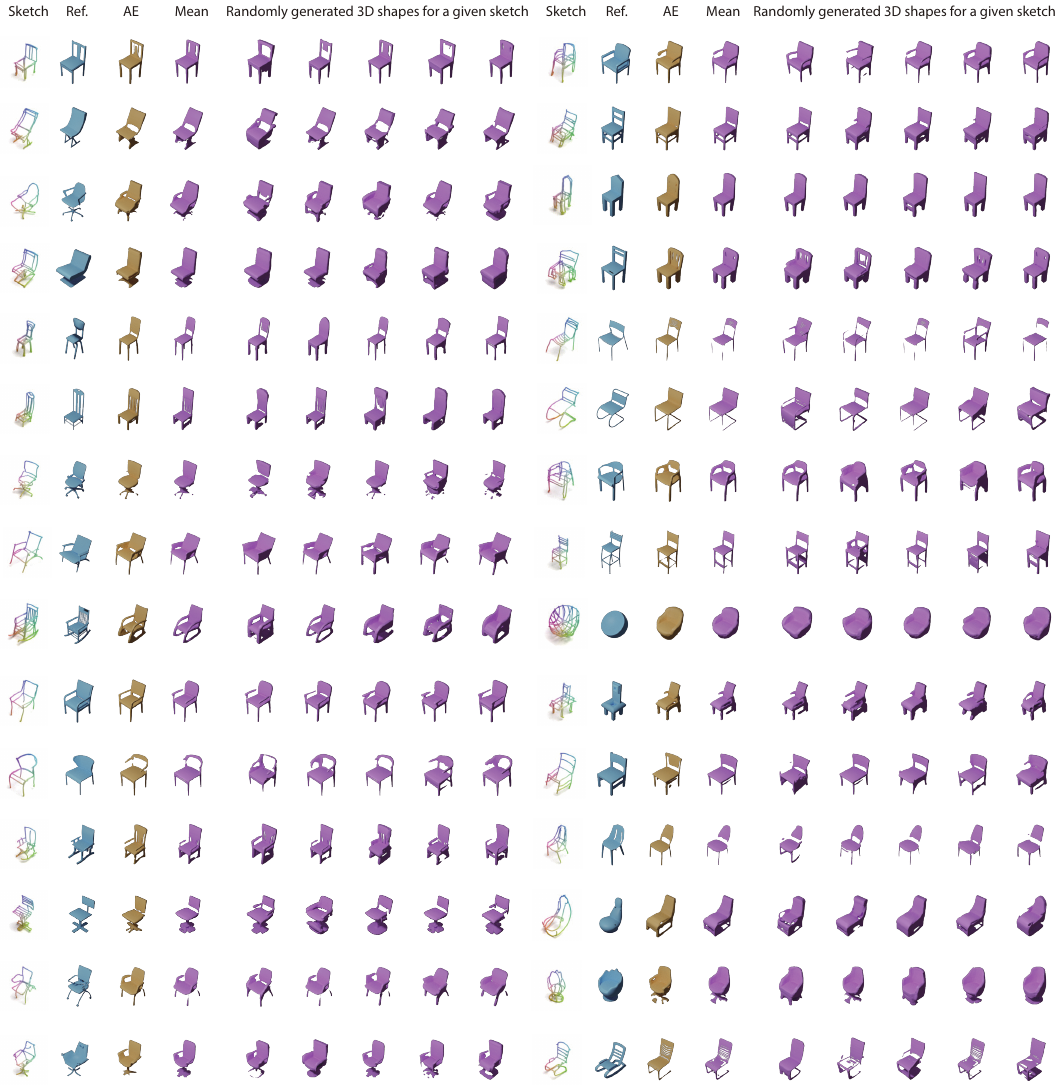}

   \caption{Additional generation results using chair sketches in {\cite{luo2021fine}}. `Ref.' shows the reference 3D shape. `AE' shows the deterministic prediction by our AE from the first stage of our method. `Mean' denotes the shape reconstructed from the sample corresponding to the mean of the conditional distribution. And finally, we show 5 randomly generated shapes conditioned on the input sketch, sorted in the order of fidelity to a reference shape.
    }
   \label{fig:more_results}
\end{figure*}

More generation results of the proposed method are shown in \cref{fig:more_results}. Please note that the scales of sketches and 3D shapes differ in these visual results due to the fact that we use different visualization tools. 
To demonstrate that sketches and reconstructed shapes generally are well-aligned, we provide \cref{fig:alignment}, created using MeshLab's snapshot.

\subsection{Performance analysis as a function of sketch accuracy}

We use a small VR sketch dataset proposed in \cite{luo2020towards} as an additional test set to explore how our method generalizes to sketches of varying quality. 
This dataset contains Freehand Sketches (FS) of 139 chairs and 28 bathtub shapes from the ModelNet10 test set.
% \todo{rename HS to FS}
% collected from novices who are asked to sketch on the surface of shapes. 
It also contains corresponding curve networks (CNs), which are the minimal set of curves required to accurately represent a 3D shape \cite{gori2017flowrep}.
Therefore, our method is expected to produce 3D reconstructions from these networks with lower diversity and better sketch fidelity than from freehand sketches. 

\begin{figure*}[h]
  \centering
   \includegraphics[width=\linewidth]{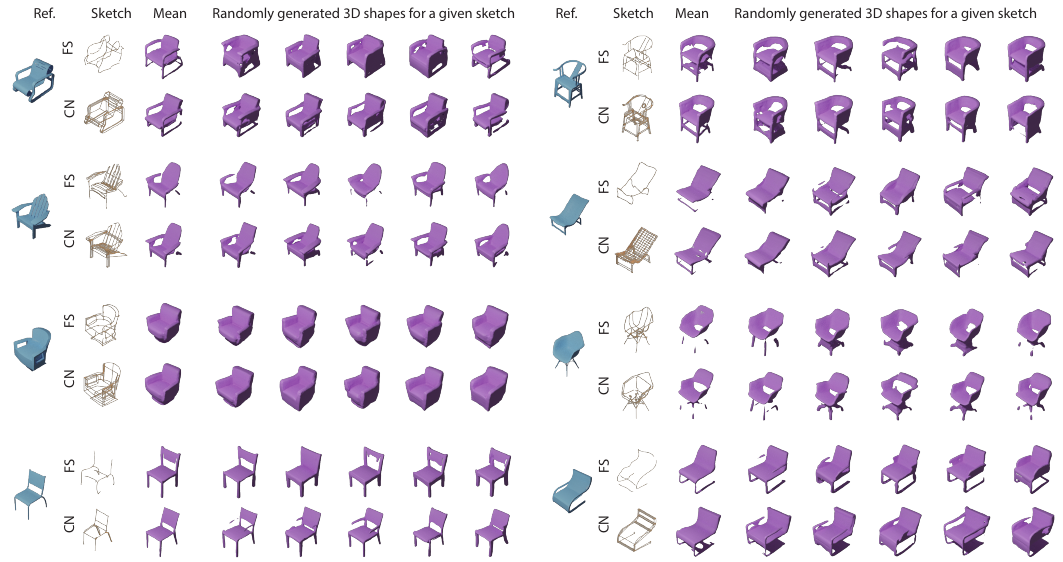}

   \caption{Generation results using chair sketches from {\cite{luo2020towards}}. `Ref.' shows the reference 3D shape. `Mean' denotes the shape reconstructed from the sample corresponding to the mean of the conditional distribution. And finally, we  show 5 randomly generated shapes conditioned on the input sketch.
    }
   \label{fig:chair}
\end{figure*}

\paragraph{Abstract sketches vs.~clean curve networks}
We show a qualitative comparison of reconstruction results from abstract sketches and matching detailed curve networks in \cref{fig:chair}. It can be observed that, indeed, the sampled shapes conditioned on the curve networks contain less diversity. 

We provide a numerical comparison between generations conditioned on sketches and curve networks of the 139 chair instances in \cref{tab:FS_vs_CN}. 

If we compare the metric on the curved networks and freehand sketches from \cite{luo2020towards} for 3D shapes from the ModelNet dataset, then we can see that indeed the reconstructions from clean curve networks are much closer to the reference 3D shape ($\mathcal{F}_{\text{shape}}$), the input ($\mathcal{F}_{\text{sketch}}$) (in the case of a curve network ``sketch" = ``curve network"), and the diversity of generated shapes is smaller ($\mathcal{D}_{\text{gnrtns}}$).

We can also see that for the two test sets of freehand sketches: the one used in the main paper and the one from \cite{luo2020towards}, the diversity and similarity to a reference 3D shape are both similar. 
However, the fidelity to the sketch is larger, which we attribute to a large number of sparse sketches in that test set, as can be observed by comparing sketch samples in \cref{fig:chair} and \cref{fig:more_results}.

\emph{Overall, this experiment shows good generalization properties to sketch styles and shapes from a different dataset of the same category.}
% Accurate sketches like curve networks will result in better fidelity to ground truth and to sketch itself, but reduce diversity at the same time. \todo{Make an analysis of sketch fidelity, shape fidelity, and diversity.} The lower fidelity to sketch on {\cite{luo2020towards}} human sketches may be due to the larger range of variation in sketch quality as exhibited in \cref{fig:chair}.

\begin{table}[t]\centering
\resizebox{\columnwidth}{!}{%
\begin{tabular}{l|c|ccc}\toprule
Data &$\mathcal{F}_{shape}(G^{{e}^c}) \downarrow$ &$\mathcal{F}^{avg}_{\text{sketch}}(G^{d^c}) \downarrow$ &$\mathcal{F}^{\text{avg}}_{\text{shape}}(G^{d^c}) \downarrow$ &$\mathcal{D}_{\text{gnrtns}} \uparrow$ \\ \midrule
{\cite{luo2021fine}} &0.368 &{0.017}$\pm$0.034 &0.431 &{0.161} \\ \midrule
{\cite{luo2020towards}} FS &0.366 &0.028$\pm$0.036 &0.412 &0.161 \\
{\cite{luo2020towards}} CN & 0.237 &0.017$\pm$0.032 &0.298 &0.142 \\
\bottomrule
\end{tabular}
}
\vspace{1pt}
\caption{Numerical comparison between generations results from freehand sketches (FS) and curve networks (CN). 
The first line shows the results from the main paper where the test set of freehand sketches represents shapes from the ShapeNetCore dataset. 
Note that the training data comes from the same dataset, but a different set of shapes. 
The second line shows the results from the freehand sketches from {\cite{luo2020towards}}, which represent shapes from the ModelNet dataset.
Finally, the third line shows the generation results on the curve networks for the shapes from the ModelNet dataset.
}
\label{tab:FS_vs_CN}
\end{table}

\begin{figure*}[h]
  \centering
   \includegraphics[width=\linewidth]{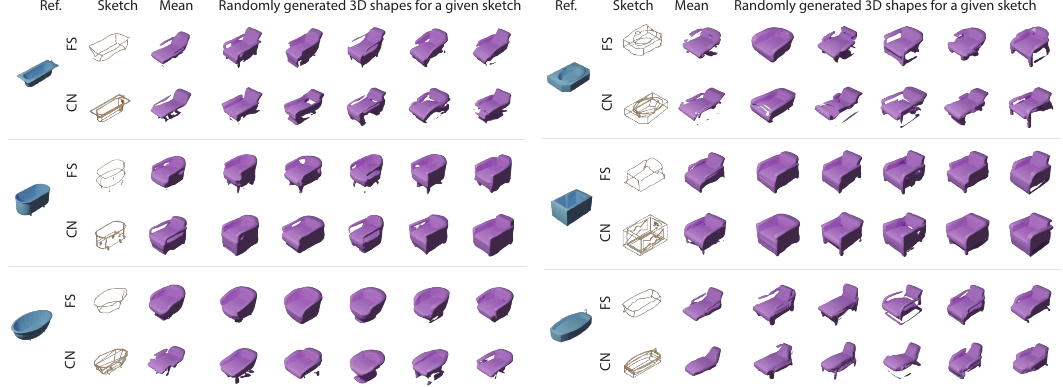}

   \caption{Generation results using bathtub sketches from {\cite{luo2020towards}}. `Ref.' shows the reference 3D shape. `Mean' denotes the shape reconstructed from the sample corresponding to the mean of the conditional distribution. And finally, we  show 5 randomly generated shapes conditioned on the input sketch.
    }
   \label{fig:bathtub}
\end{figure*}

In \cref{fig:bathtub}, we show what happens if the test set comes from a completely different distribution than the training set. Namely, we use as conditioning freehand sketches and curve networks of bathtubs. 
\emph{It can be observed, that our approach produces valid or physically plausible chairs which follow the overall shape of the conditioning, in turn demonstrating the robustness of our method, and its ability to generate new shapes.}

\subsection{Comparison with retrieval}

We compute our fidelity and diversity metrics on the top 5 retrieval results of \cite{luo2022structure} with the ground-truth shape in the gallery. \cref{tab:num_res} shows that retrieval results have much lower fidelity to sketches and ground-truth shapes. Additionally, note that if there is no shape close to the sketch in the 3D gallery, the retrieval has no chance to succeed.

\subsection{Evaluation on sketches with style not seen during training}
Following \cite{luo2020towards}, our test set includes 5 participants those sketches are not used for training. 
\cref{tab:num_res} shows generalization to lower quality 202 VR chair sketches from memory \emph{FVRS-M} \cite{luo2022structure}.

\subsection{Comparison with 2D sketch-based modeling}
We compare with the performance of a sketch-based modeling method, following closely ours, but conditioned on 2D sketches by novices \cite{qi2021toward}. 
These sketches match 3D shapes in our training and test sets of VR sketches. 
We feed all 3 available views to the \cite{he2019view} encoder, which we train with $\mathcal{L}_{\text{align}}$ (Eq.12) and $\mathcal{L}_{\text{SDF}}$ (Eq.5) and leverage our pre-trained decoder, similar to the setting in \cref{tab:encoder}, lines 3-4, in the main paper.  
We obtain $\mathcal{F}_{\text{shape}}^{2D}=0.833$ which is much worse than on 3D sketches: $0.437$ (\cref{tab:encoder}, line 3).

\begin{table}[t]\centering
\vspace{-8pt}
\resizebox{\columnwidth}{!}{%
\footnotesize{
\begin{tabular}{l|ccc}\toprule
Experiment &$\mathcal{F}^{\text{avg}}_{\text{sketch}}(\cdot) \downarrow$ &$\mathcal{F}^{\text{avg}}_{\text{shape}}(\cdot) \downarrow$ &$\mathcal{D}_{\text{gnrtns}} \uparrow$ \\ \hline
Unseen 5 &0.019$\pm$0.035 &0.460 &0.171 \\ 
FVRS-M  &0.030$\pm$0.041 &0.431 &0.161 \\ 
Retrieval {\cite{luo2022structure}} &0.061$\pm$0.056 &1.132 &1.138 \\ 
\bottomrule
\end{tabular}
}
}
\vspace{-5pt}
\caption{The first two rows show the performance of our method on (1) the set of sketches by 5 participants those sketches were not observed during training -- they represent new shapes and new styles;
(2) sketches from the FVRS-M dataset  \cite{luo2022structure}.
The last row provides numerical evaluation on the quality of the top-5 3D shapes returned by the retrieval method by Luo et al.~\cite{luo2022structure}.}
\label{tab:num_res}
\end{table}

\subsection{Interpolation in sampling space}
\begin{figure*}[h]
  \centering
   \includegraphics[width=\linewidth]{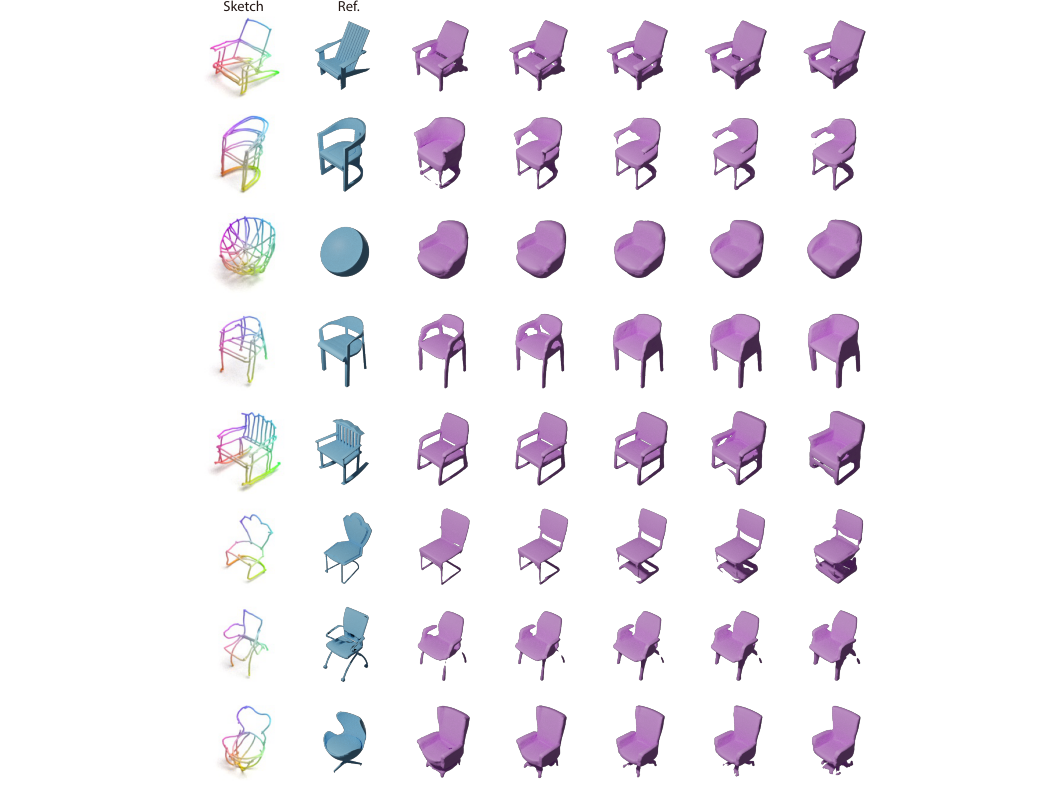}

   \caption{Interpolation results in sampling space. Note the smooth transitions between the first and last purple objects.
    }
   \label{fig:interp}
\end{figure*}

To verify the continuity of our sampling space, we perform interpolation between two generated shapes in the latent space of CNF. 
Given the same sketch condition, we first generate two random samples. 
We then linearly interpolate their latent codes to create 3 additional samples.
We then pass those new codes through the reverse path of the CNF with the same sketch condition to map back to the embedding space of the autoencoder, which allows us to compute their SDF representation. The visual interpolation results are shown in \cref{fig:interp}.

% The results indicate that our method shows potential to fine-tune and manipulate the generated results at a more detailed level.
The figure shows that the interpolation is quite smooth.
Therefore, the user can choose any two generated results and explore the shapes in between. \emph{This further supports shape exploration enabled by our method.}

\section{Training details}

\subsection{Sampling SDF for training}
As we mention in \cref{sec:implementation_details}, to prepare the training data, we compute SDF values on two sets of points: close to the shape surface and uniformly sampled in the unit box, following \cite{park2019deepsdf}.

\paragraph{Close to the shape surface} 
First, we randomly sample 250k points on the surface of the mesh.
During sampling, the probability of a triangle to be sampled is weighted according to its area.
We then generate two sets of points near the shape surface by perturbing each surface point coordinates with random noise vectors sampled from the Gaussian distribution with zero mean and variance values of 0.012 or 0.035. 

To obtain two spatial samples per surface point, we perturbed each surface point along all three axes (X, Y, and Z) using zero mean Gaussian noise with variance values of 0.012 and 0.035. 

\paragraph{Uniformly sampled in the unit box} 
Second, we uniformly sample 25k points within the unit box. This results in a total of 525K points being sampled. 

\paragraph{Loss evaluation} 
When computing the SDF $L_1$ loss (\cref{eq:SDF_loss}) during training, we sample a subset of 8,192 points from the pre-computed set of points for each SDF sample. 
We then compute the $L_1$ loss between the predicted SDF values and the ground truth SDF values at the sampled 3D points coordinates.

\subsection{Sampling point clouds from sketch strokes and 3D shapes}

To obtain point cloud representations of shapes and sketches, prior to training, we sample 4,096 points for both shapes and sketches.
For shapes, we uniformly sample from the mesh surface\footnote{\url{https://github.com/fwilliams/point-cloud-utils}}. 
VR sketches are made up of strokes that include a set of vertices and edges.
Luo et al.~\cite{luo2021fine} provide sketch point clouds, each containing 15,000 points sampled uniformly from all sketch strokes. 
We apply Furthest Point Sampling (FPS) to sample 4,096 points from the 15,000 points.

\end{document}